\documentclass[conference]{IEEEtran}
\IEEEoverridecommandlockouts
\usepackage{cite}
\usepackage{amsmath,amssymb,amsfonts}
\usepackage{algorithmic}
\usepackage{graphicx}
\usepackage{hyperref}
\usepackage{textcomp}
\usepackage{multirow}
\usepackage{booktabs}
\usepackage{subcaption}
\usepackage{xcolor}
\def\BibTeX{{\rm B\kern-.05em{\sc i\kern-.025em b}\kern-.08em
    T\kern-.1667em\lower.7ex\hbox{E}\kern-.125emX}}
\hypersetup{
     colorlinks = true,
     linkcolor = blue,
     anchorcolor = blue,
     citecolor = green,
     filecolor = magenta,
     urlcolor = cyan
     }
\begin{document}

\title{\vspace{+1.0cm}SSTN: Self-Supervised Domain Adaptation Thermal Object Detection for Autonomous Driving\\
}

\author{\IEEEauthorblockN{Farzeen Munir, Shoaib Azam and Moongu Jeon}
\IEEEauthorblockA{School of Electrical Engineering and Computer Science\\
Gwangju Institute of Science and Technology\\
Gwangju, South Korea\\
Email: {(farzeen.munir,shoaibazam, mgjeon)@gist.ac.kr}
}
}

\maketitle

\begin{abstract}
The perception of the environment plays a decisive role in the safe and secure operation of autonomous vehicles. The perception of the surrounding is way similar to human vision. The human's brain perceives the environment by utilizing different sensory channels and develop a view-invariant representation model. In this context, different exteroceptive sensors like cameras, Lidar, are deployed on the autonomous vehicle to perceive the environment.  These sensors have illustrated their benefit in the visible spectrum domain yet in the adverse weather conditions; for instance, they have limited operational capability at night, leading to fatal accidents. This work explores thermal object detection to model a view-invariant model representation by employing the self-supervised contrastive learning approach. We have proposed a deep neural network Self Supervised Thermal Network (SSTN) for learning the feature embedding to maximize the information between visible and infrared spectrum domain by contrastive learning. Later, these learned feature representations are employed for thermal object detection using a multi-scale encoder-decoder transformer network. The proposed method is extensively evaluated on the two publicly available datasets: the FLIR-ADAS dataset and the KAIST Multi-Spectral dataset. The experimental results illustrate the efficacy of the proposed method. 
\end{abstract}


\section{Introduction}
The recent technological advancements have enabled autonomous vehicles to be deployed on the road, ensuring the safety standard as indicated by SOTIF-ISO/PAS-21448\footnote{https://www.daimler.com/innovation/case/autonomous/safety-first-for-automated-driving-2.htm}.  Keeping in this context, the perception of the autonomous vehicle has an integral element in defining the environment for the autonomous vehicle \cite{rosique2019systematic}. For the environment's perception, the most common exteroceptive sensors include cameras, Lidar, and radar. These sensors have their own merits and demerits; for instance, cameras (visible spectrum) provide an in-depth high resolution of the environment but have an illumination problem and cannot be utilized in night conditions. Lidar uses laser light for modelling the environment and provides the 3D point-cloud data of the environment. Despite providing the 3D information of the environment, Lidar is extremely expensive and has a resolution problem in adverse weather conditions. Radars also enable the autonomous vehicle to identify small objects near the vehicle but have a low resolution at the range. Therefore, these sensors have limited operating capability in perceiving the environment for the autonomous vehicle at night\cite{van2018autonomous}. The utilization of a thermal camera in the autonomous vehicle's sensor suite provides the necessary solution for the perception of the environment at night yet requires an efficient perception algorithm in terms of object detection\cite{leira2021object}.
\begin{figure}[t]
      \centering
      \includegraphics[width=7cm]{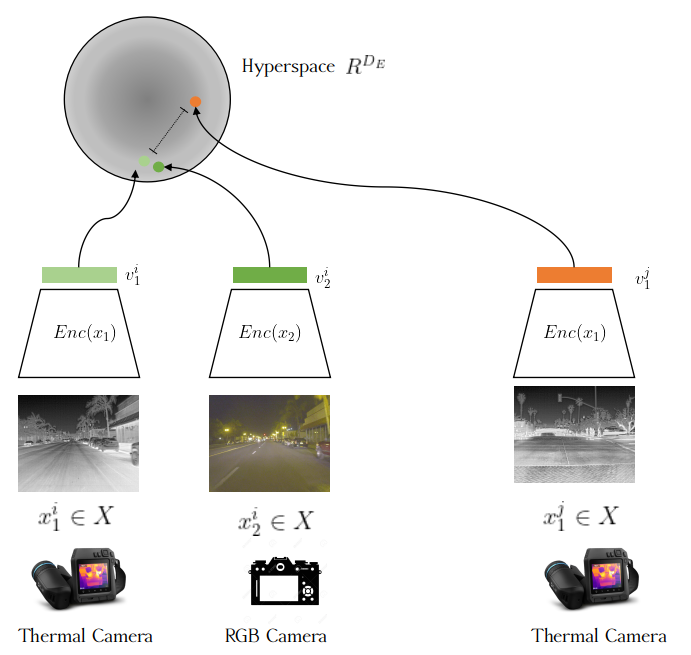}
      \caption{Given a set of multi-sensor data, self-supervised contrastive learning learn representation to bring same scene together in embedding space (hyperspace), while pushing different scene apart .}
      \label{concept}
\end{figure}
\par 
Object detection is an essential component that forms the basis for the perception of the autonomous vehicle. Although the use of the deep neural network has illustrated the significant improvement in object detection, yet most of the efforts have focused on the visible spectrum images that include YOLO\cite{yolov3}, SSD\cite{ssd}, Faster-RCNN\cite{faster}, RefineDet\cite{refine}, M2Det\cite{m2det}. There is still room for improvement in the thermal object detection domain compared to its counterpart, visible spectrum object detection. Some of the literature focuses on thermal object detection by employing domain adaptation techniques to transfer the knowledge from the visible spectrum to the thermal domain for object detection in the thermal domain. In addition, the visible spectrum domain object detectors have also been used on thermal images for object detection. All these approaches for thermal object detection have utilized the supervised learning mechanism for feature representation. The feature representation by employing the self-supervised learning techniques using the unlabelled dataset provides an alternative solution to its surrogate supervised learning approach.
\par
Learning good representation without human supervision is a fundamental problem in computer vision. The two main classes that illustrate the possible solution are generative and discriminative approaches. Generative approaches in the form of auto-encoder\cite{salakhutdinov2009deep} and generative adversarial network \cite{goodfellow2014generative} model the pixels in the input space that leads to the computationally expensive representation learning for feature extraction\cite{hinton2006fast} \cite{kingma2013auto}. Besides, the discriminative approaches utilize the objective function for learning the representation by designing a pretext task to train the deep neural network\cite{gidaris2018unsupervised}. This framework of learning the representation using the pretext task confide in the heuristics of designing the pretext task that limits the generalization of learned representations. The discriminative approach in the form of contrastive learning is utilized to learn the latent representation to overcome the heuristics of pretext tasks\cite{chen2020simple}\cite{oord2018representation}.
\par 
This work relies on the hypothesis that the view-invariant representations are encoded by the human brain \cite{den2012prediction} \cite{hohwy2013predictive}. Humans view the surrounding environment with different sensory modalities. These sensory modalities are incomplete and noisy, but the prominent factor about the environment, for instance, geometry, semantics, and physics, are shared between these sensory modalities, illustrate the powerful model representation invariant to different views \cite{smith2005development}. This study explores this hypothesis for the learning representations that capture information shared between the visible and thermal domains. For this purpose, we have employed the contrastive learning approach to learn the features embedding in the latent space projecting the distance (generally measured as Euclidean distance) to nearby points for the same scene corresponds to two sensory domains and far apart in the context of different views.  Fig. \ref{concept} illustrates the pictorial overview of our proposed framework. The proposed neural network SSTN is designed in a two-stage network. The first stage, self-supervised contrastive learning, corresponds to the feature representation by contrastive learning and maximizing the mutual information between the two domains and transferring the domain knowledge of visible spectrum to the thermal domain in a self-supervised manner for the object detection in the thermal domain. The later stage illustrates the thermal object detector's design consisting of a multi-scale encoder-decoder transformer network architecture that incorporates the features embedding from the self-supervised contrastive learning stage. The increment of the mean average score of the proposed method with ResNet101 in comparison to other state-of-art methods is 2.57\% for FLIR-ADAS and 2.37\% for the KAIST Multi-Spectral dataset.
\par 
The main contributions of our work are:
\begin{enumerate}
    \item We have designed a self-supervised domain adaptation framework (SSTN) for the thermal object detection. The proposed SSTN network have illustrated the utilization of a self-supervised contrastive learning approach to maximize the information between visible and thermal domain for the object detection in the thermal domain.
    \item The contrastive learning approach is incorporated to cater to the scarcity of labelled dataset and has learned the feature representation in a self-supervised manner. In addition, the proposed work, to the best of our knowledge, is the first work to incorporate self-supervised contrastive learning in multi-sensor framework.  
    \item Further, we have extended the feature embedding learned in a self-supervised manner to a multi-scale encoder-decoder transformer network for thermal object detection. 
\end{enumerate}

The rest of the paper is organized as follows: Section II gives overview of related work.  Section III explains the proposed methodology. The experimentation and results are discussed in Section IV, and section V concludes the paper.
\begin{figure*}[t]
      \centering
      \includegraphics[width=14cm]{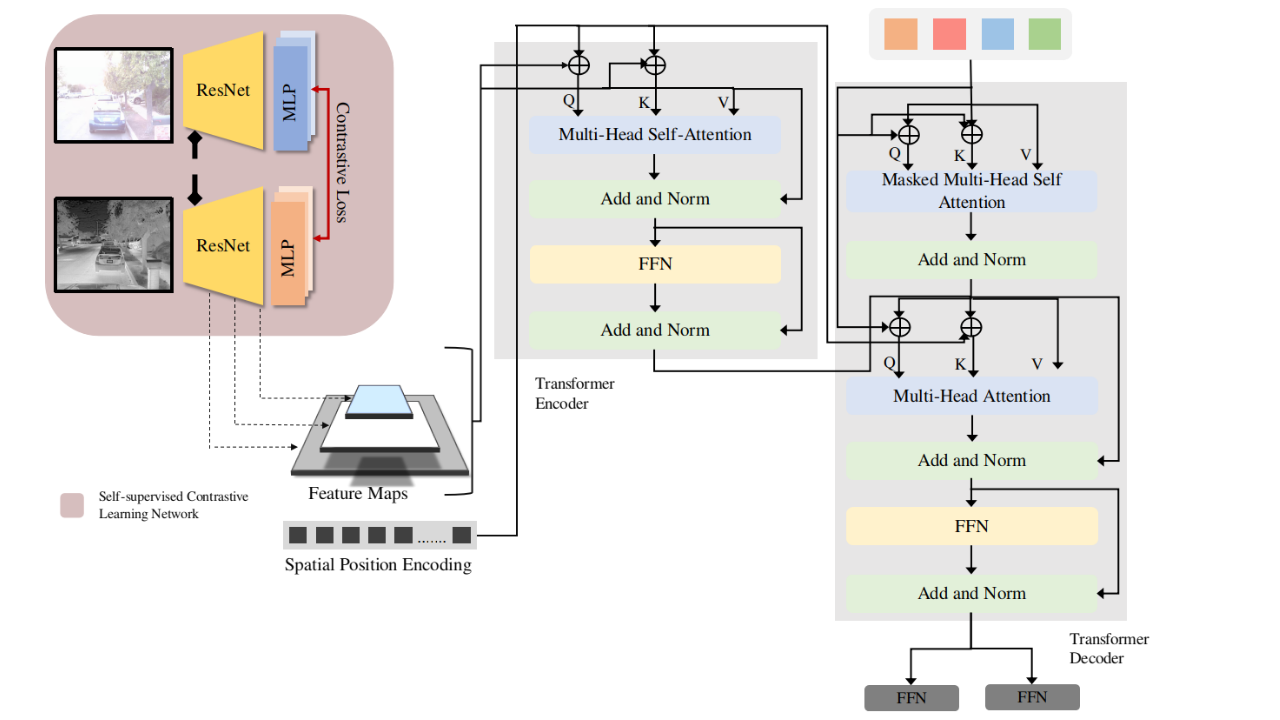}
      \caption{The overall framework for the proposed method. It includes a self-supervised contrastive learning network and multi-scale feature embedding for the transformer encoder-decoder network for the thermal object detection.}
      \label{framework}
\end{figure*}
\section{Related Work}
Deep neural networks have been used as a function approximator for predicting and classifying objects in the visible and infrared spectrum domain. In literature, extensive research has carried out on designing the robust object detector for the visible spectrum domain\cite{yolov3}\cite{m2det}\cite{faster}\cite{ssd}. For thermal object detection, the research focuses on employing feature engineering by fusing the information.\cite{krivsto2020thermal} have used visible spectrum domain object detector YOLOv3 for the person detection in the thermal domain and have benchmarked the performance of YOLOv3 with Faster-RCNN, SSD, and Cascade R-CNN. \cite{ghose2019pedestrian} illustrates the thermal object detection problem by augmenting the thermal image frame with the corresponding saliency maps to provide an attention mechanism for pedestrian detection in the thermal domain using Faster-RCNN. The multispectral images (thermal, near-infrared (NIR)) are fused and used to train the YOLOv3 for thermal object detection. Besides deep neural network, classical image processing approaches are also applied for thermal object detection \cite{soundrapandiyan2015adaptive}. \cite{baek2017efficient} \cite{li2012effective} have applied the HOG and local binary pattern for feature extraction and trained the support vector machine (SVM) classifiers for the object detection in the thermal domain. Much research has also been done by incorporating the domain knowledge transfer between the visible and infrared domain. \cite{devaguptapu2019borrow} have employed the generative adversarial approach for modelling the thermal image from the visible RGB image. This generated thermal image is utilized for training the variant of Faster-RCNN to detect the object in both RGB and thermal images. Similarly, a cross-domain semi-supervised learning framework is proposed for feature representation in the target domain\cite{yu2019unsupervised}. In addition, researchers have also illustrated the use case of transfer learning between domain using generative adversarial networks \cite{zheng2020p} \cite{royer2020xgan}.
\par

Self-supervised learning provides a way to learn the features embedding without explicit availability of labelled dataset. This representation learning in the context of image domain uses predictive approaches to learn the feature embedding by devising the artificially handcrafted pretext tasks \cite{doersch2015unsupervised} \cite{noroozi2016unsupervised} \cite{zhang2016colorful} \cite {chen2019self}. Although these pretext tasks illustrate the promising feature representations, yet these approaches depend on the heuristics of designing the pretext tasks. Another approach to learn feature embedding is to use the contrastive learning method. \cite{hadsell2006dimensionality} have proposed an approach to learn the representation by contrasting the negative and positive pairs. Similarly, to improve the computation efficacy, a method employing the memory bank to store the instance class representation is proposed in this study \cite{wu2018unsupervised}. This work is further investigated, and the memory bank method is replaced by in-batch negative sampled sampling \cite{ji2019invariant} \cite{ye2019unsupervised}. In addition, \cite{chen2020simple} have proposed a simple framework to learn the feature representation through contrastive learning irrespective of memory bank or specialized architectures. This framework explores data augmentation techniques, a learnable nonlinear transformation layer between representation and contrastive, representation learning with contrastive cross-entropy loss. \cite{tian2019contrastive} have explored the contrastive predictive coding and proposed contrastive multiview coding for the representation learning between the different view of the same scene by augmenting them in color channel space. Similarly, a supervised contrastive loss is introduced in this work \cite{khosla2020supervised}.

\section{Self-Supervised Domain Adaptation Thermal Object Detection for Autonomous Driving}
Learning effective visual representation without labelled data is a challenging problem. The goal is to learn representations that capture useful features for deep-learning-based object detection.  This work explicitly explores the co-occurrence of information in multi-sensor data obtained from the thermal image and RGB image.  A two-stage deep neural network,  Self-supervised thermal network (SSTN), is proposed, where the first stage consists of pre-training involving self-supervised representation learning by contrastive learning approach. The second stage consists of a multi-scale encoder-decoder transformer network for object detection in the thermal domain based on features learned in the previous stage.
\subsection{Self-supervised Contrastive Learning}
The thermal camera captures the surrounding information using heat signatures exhibiting from different objects. These heat signatures are represented in the form of images for better representation learning. To do so, we have adopted multi-spectrum self-supervised contrastive learning for feature representation.
Fig. \ref{framework} shows the self-supervised contrastive learning network. Suppose the set $X$ consists of thermal and visible RGB images of the same scene represented as $x_1$ and $x_2$. A stochastic data augmentation module transforms the input $X$ into two correlated images, which share information of the same scene.  
A neural network encoder  maps $X$ to representation vector, $v=Enc(X) \in R^{D_E} $, and $R^{D_E} $ normalized to unit hyperspace in $R^
{D_E}$ \cite{khosla2020supervised},\cite{chen2020simple}. Both $x_1$ and $x_2$ are separately fed to the same encoder network to obtain a pair of representation vectors. The ResNet network is adopted as an encoder network \cite{a10}, due to its applicability and common usage. We instantiate a multi-layer perceptron as a projection network that maps $v$ to a vector $w=Proj(v)\in R^{D_E}$ \cite{a3}. The multi-layer perceptron consists of a single hidden layer with a size of $2048$ and an output layer of size $F_p=128$. The output vector is also normalized to lie in the unit hypersphere, facilitating the calculation of the inner product in projection space \cite{tian2019contrastive}.
\par
The contrastive loss is formulated for multi-spectrum image batch. For a set of $N$ random pairs of thermal and RGB images $\{X_k\}{k=1....N},$ are selected. The corresponding batch size $\{X_l\}{l=1....2N}$ for this configuration is $2N$, which is fed for training the network.
$x'_{2k}$ and $x'_{2k-1}$ represent thermal and RGB sample pair respectively of $X_k$. In a multi-spectrum batch, $i \in I \equiv \{ 1...2N \}$ is an index of the randomly selected thermal sample, and $j(i)$ is an index of its corresponding RGB sample. The Eq.\ref{equ2} gives the self-supervised contrastive learning loss function \cite{tian2019contrastive}\cite{a5}. 
 \begin{align}
 \label{equ2}
   L^{self}=\sum_{i \in I} L_{i}^{self}=-\sum_{i \in I}log\frac{exp(w_i\cdot w_{j(i)}/\tau)}{\sum_{b \in B(i)}exp(w_i\cdot w_b/\tau)},
\end{align}
Here, $w_l=proj(Enc(X'_l)) \in R^{D_E}$, $\tau \in R^+$ is a scalar parameter for temperature, $\cdot$ represents the inner product of the two vectors, and $B(i) \equiv I \setminus \{i\}$. $i$ represent index of the anchor and $j(i)$ is called a positive in the batch, while other $2(N-1)$ indexes $(\{k\in B(i) \setminus \{j(i)\}\})$ are negatives. To clarify each anchor $i$ have one positive pair and $2(N-1)$ negative pairs.
\par 
We have used the self-supervised contrastive learning stage as a backbone to the transformer encoder-decoder. The feature map learned through the contrastive learning stage are input to the transformer encoder as a set query and key elements. 
\subsection{Encoder-Decoder Transformer Network}
\subsubsection{Transformer Encoder}
A transformer encoder inputs low-resolution feature maps which is obtained from convolution neural network backbone . In this work, multi-scale feature maps $\{f^l\}_{l=1}^{L-1}$ are obtained from the self-supervised contrastive learning stage, where $f^l \in \mathbb{R}^{C\times H_l \times W_l}$. The multi-scale feature maps $\{f^l\}_{l=1}^{L-1}$, where $L=4$ are extracted from pre-training stages $C_3$ to $C_5$ \footnote{The convolution blocks as explained in \cite{a10}} in the ResNet encoder \cite{a10}. The feature maps are transformed by $1 \times 1$ convolution  $\tilde{f} \in \mathbb{R}^{C \times HW}$. The $C_5$ feature map is obtained by performing $3 \times 3$ and stride $2$ convolution. All multi-scale feature maps are of $C=256$ channels. The encoder inputs and outputs the same resolution feature maps. Each transformer encoder layer consists of a multi-scale, multi-head self-attention module \cite{a6}\cite{a7}. The transformer encoder inputs key and query elements, $\varsigma_q$ and $\varsigma_k $ denotes a set of query and key elements. Let $q \in \varsigma_q$ represent a query element with representation features $z_q \in \mathbb{R}^C$ and $k \in \varsigma_k $ represents a key element with feature embedding $f_k \in \mathbb{R}^C$. The multi-head, multi-scale self-attention features are calculated by 
 \begin{align}
 \label{equ3}
   (z_q,\{f^l\}_{l=1}^L)=\sum_{m=1}^{} W_m[\sum_{l=1}^{L}\sum_{k \in \varsigma_k} A_{mlqk}\cdot W'_mf^l],
\end{align}
where $m$ represents the number of attention head, $l$ shows the input feature maps levels. $W_m \in \mathbb{R}^{C_v \times C}$ and $W'm \in \mathbb{R}^{C \times C_v}$ represent weight tensor which are learned during the training. $(C_v=C/M)$. Attention weight tensor $Amlqk 	\propto exp\{\frac{z_q^TU_m^TV_mf_k}{\sqrt{C_v}}\}$ is normalized by $\sum_{l=1}^L\sum_{k \in \varsigma_k} Amlqk=1$, here $U_m$ and $V_m$ are learnable weight tensors. The spatial position encoding are learned and shared among all the attention layers of an encoder for a certain query, key and feature map pair \cite{a8}.
\subsubsection{Transformer Decoder}
The Transformer decoder mimics the encoder framework of sub-layers. The layers consist of multi-scale cross-attention and self-attention modules. Each type of attention module input object queries. The object queries are initially set to zero, and $N$ object queries are learnt positional encodings and encoder memory. In the case of the cross-attention module, object queries derive features from multi-scale feature maps, and output feature maps from the encoder are the key elements.  Moreover, in the self-attention module, attention is computed amid object queries. The decoder transforms the $N$ object queries into output embeddings. The output embeddings are then fed to the feed-forward network, which computes the bounding box coordinates and class labels. 
\subsubsection{Feed-Forward Network}
The decoder's output embedding is given to the feed-forward network, which consists of 3 layer perceptron network with RELU activation and a hidden dimension of $k$. A fully connected layer is used as the final layer, which performs the linear projection to predict the output. The output consists of class labels and the bounding box coordinates comprising height, width and the normalized centre coordinates. Moreover, a no-class label is also predicted for the images with no object present in the image.  

\section{Experimentation and Results}
\subsection{Dataset}
In this study, we have used two publicly available datasets i.e. FLIR-ADAS dataset\footnote{https://www.flir.in/oem/adas/adas-dataset-form/}, and the KAIST Multi-Spectral dataset \cite{kiast}.  FLIR-ADAS dataset contains 9214 thermal and RGB image pair, and objects are annotated in MS-CoCo format, i.e. a bounding box and class labels. The aforementioned dataset has 3 class labels, car, person and bicycle.  The data is collected using the FLIR Tau2 camera and each image has the resolution of $640 \times 512$. Both day and night time data is collected. We have used a standard split of the dataset in training and testing as illustrated in Table \ref{table3}.

\begin{table}[b]
\centering
\caption{FLIR-ADAS and KAIST Multi-Spectral datasets partition topology for training and testing the proposed SSTN network.}
\label{table3}
\resizebox{9cm}{!}{%
\begin{tabular}{@{}l|l|c|c@{}}
\toprule
Dataset              & Total Images & \multicolumn{1}{l|}{Train Images} & \multicolumn{1}{l}{Test Images} \\ \midrule
FLIR-ADAS            & 10228        & 8862                              & 1366                            \\
KAIST Multi-Spectral & 95000        & 76000                             & 19000                           \\ \bottomrule
\end{tabular}%
}
\end{table}
\par
The KAIST Multi-Spectral dataset has 95000 image pair of thermal and RGB image \cite{kiast}. However, in this dataset only the person class is annotated. KAIST Multi-Spectral dataset is collected using FLIR A35 camera with a resolution of $340 \times 256$ and include day and night time images. We have used the standard split of the dataset for training and testing as illustrated in Table \ref{table3}.
\subsection{Evaluation Metric}
In this study, we have used the standard MS COCO evaluation metric \cite{coco}.  Intersection over union $(IoU)$ is computed between the area covered by the ground-truth bounding box and the area covered by the predicted bounding box, given by Eq.\ref{equa12}. 
\begin{equation}
\hspace{-5cm}
    \begin{aligned}
    \label{equa12}
   IoU= \frac{A_p \cap A_{gt}}{A_p \cup A_{gt}},
\end{aligned}
\end{equation}
The True Positive (TP) is held true for the $IoU>thresh$ and False Positive (FP) is considered when $IoU<thresh$. Based on TP and FP precision and Recall are calculated shown below. 
Recall and Precision are calculated using Eq.\ref{equa13}.
\begin{equation}
   \begin{aligned}
    \label{equa13}
   & Recall =\frac{TP}{TP+FN}, Precision =\frac{TP}{TP+FP},
\end{aligned} 
\end{equation}

For each class, Average precision (AP) is computed. AP is the area under the PR curve, which is computed using the Eq.\ref{equa11}.

\begin{equation}
   \begin{aligned}
    \label{equa11}
   AP[class]=\frac{1}{\#thresh} \sum_{IoU \in thresh} AP[class,IoU],
\end{aligned} 
\end{equation}
MS COCO evaluation metric varies the threshold $(thresh)$ from $0.5$ to $0.95$ with a incremnet of $0.05$.  In this study $mAP_{IoU}=0.5$ is considered for all the experiments.
\subsection{Experimentation}
This section explains the experimentation details of the proposed network (SSTN) and discusses the results.
\subsubsection{ Self-supervised Contrastive Learning} 
The self-supervised contrastive network is trained using thermal and RGB image pair. The input images undergo augmentations that include random crop and resize, horizontal flip and colour jitters. For this study, we have only considered ResNet50 and ResNet101 as the encoder for the self-supervised network. The network is trained for $1000$ epoch with a batch size of $16$, learning rate of $0.05$ and temperature of $0.07$.  The network is implemented using Pytorch deep learning library, and the weights of the network are optimized using the loss function given by Eq.\ref{equ2}. All the experimentation is performed on three GPU machine, where each GPU has a memory of 12Gb.
\subsubsection{Faster-RCNN Baseline}
The faster-RCNN with ResNet backbone is employed to develop a baseline for the proposed network \cite{faster}. Two types of experiments are performed. First, the network is trained in a supervised manner on thermal image data with no pre-training weights. Second, the network backbone is initialized by the features optimized by the self-supervised contrastive network and finetune on labelled thermal image data. We have adopted Pytorch implementation of Faster-RCNN for this purpose. The input data is augmented with a random scale, randomly crop and resize and horizontally flipped. It is noted that input image augmentation is kept the same for all the experiments conducted with Faster-RCNN and multi-scale encoder-decoder transformer network. The network is trained with a stochastic gradient descent optimizer with a learning rate of $0.02$
\subsubsection{ Encoder-Decoder Transformer Network}
 The transformer encoder-decoder network is trained using an AdamW \cite{a9} optimizer with an initial learning rate of $2e^{-4}$, the backbone learning rate of $2e^{-5}$ and weight decay of $1e^-4$. The batch size is set to 2, and the network is trained for 100 epochs.  The multi-scale feature maps are extracted from a self-supervised contrastive learning network, as shown in Fig. \ref{framework}. For the transformer number of encoder and decoder are equal to $6$. Absolute position encodings are considered a function of sine and cosine at different frequencies and then concatenated to achieve the final position encoding across the $C$ channels. It is to be noted that in all the experimentation, the hyper-parameters values are determined heuristically.
\par
An optimal bipartite matching scheme is selected for loss calculation inspired by \cite{a8}. The proposed method outputs a set of $N$ predictions. Eq.\ref{equ6} defines the bipartite matching loss between a set of ground-truth $(y)$ and a set of the predicted labels $\hat{y}$.$S_N$ illustrates the permutations of N elements.  $L_{match}(y_j,\hat{y}_{\eta(j)})$ corresponds to pair-wise matching of predicted and ground-truth labels. A Hungarian algorithm is used to compute assignments between ground-truth and predictions.   The matching cost aggregates both the class labels and the bounding box between the predictions and the ground-truth. Lets element $j$ indexes the ground-truth set $y_j = (c_j,b_j)$, where $c_j$ is the class labels and  $b_j \in [0,1]^4$ is normalized bounding box vector including center coordinates, width and height.  The predicted class probability is define by  $\hat{p}_{\eta{j}}(c_j)$ and bounding box is given by $\hat{b}_{\eta{j}}$. To match the prediction and ground-truth set, direct one-to-one correspondence is found without any duplicates. The Hungarian loss function is illustrated by Eq.\ref{equ7}.
\begin{align}
\hspace{-2cm}
    \label{equ6}
    \tilde{\eta} = \underset{\eta \in S_N}{argmin} \sum_{j}^{N}L_{match}(y_j,\hat{y}_{\eta(j)}),
\end{align}
$S_N$ illustrates the permutations of N elements.  $L_{match}(y_j,\hat{y}_{\eta(j)})$ corresponds to pair-wise matching of predicted and ground-truth labels. A Hungarian algorithm is used to compute assignments between ground-truth and predictions.   The matching cost aggregates both the class labels and the bounding box between the predictions and the ground-truth. Lets element $j$ indexes the ground-truth set $y_j = (c_j,b_j)$, where $c_j$ is the class labels and  $b_j \in [0,1]^4$ is normalized bounding box vector including center coordinates, width and height.  The predicted class probability is define by  $\hat{p}_{\eta{j}}(c_j)$ and bounding box is given by $\hat{b}_{\eta{j}}$. To match the prediction and ground-truth set, direct one-to-one correspondence is found without any duplicates. The Hungarian loss function is illustrated by Eq.\ref{equ7}.
 \begin{flalign}
\hspace{-1cm}
    \label{equ7}
    \begin{aligned}
    L_{Hungarian}(y,\hat{y})= \sum_{j=1}^{N}[-log\hat{p}_{\eta({j})}(c_j)+ & \\ \mathbf{1}_{c_j\neq(\phi)}L_{box}(b_j,\hat{b}_{\hat{\eta}}(j))], 
    \end{aligned}
\end{flalign}
 $\hat{\eta}$ indicate the optimal assignment. The $L_{box}(.)$ computes the score for the bounding box  is represented as shown in Eq.\ref{equf}
\begin{equation}
\label{equf}
\hspace{-0.5cm}
    \begin{aligned}
    L_{box}(b_j,\hat{b}_{\hat{\eta}}(j)) = \lambda_{iou}L_{CIoU}(b_j,\hat{b}_{\hat{\eta}}(j)) &+ \\
    \lambda_{l1}\left \|b_j- \hat{b}_{\hat{\eta}}(j) \right \|,
    \end{aligned}
\end{equation}

where, $l_1$ loss with complete $IoU$ loss \cite{a10} is used to calculate $L_{CIoU}$
\begin{align}
\hspace{-0.5cm}
    \label{equb}
   L_{CIoU}=1-IoU+\frac{\rho (\mathbf{b},\boldsymbol{b_{gt}})}{c^2}+\alpha \nu, 
\end{align}
$1-IoU+\frac{\rho (\mathbf{b},\boldsymbol{b_{gt}})}{c^2}$ represents $IoU$ distance with  predicted bounding box $b$ and ground-truth bounding box $b_gt$. $c$ is the digonal length of smallest enlosing that covers the two boxes. $\rho$ shows the euclidean distance. $\alpha$ is a positive trade-off parameter and $\nu$ illustrates the consistency of aspect ratio given by Eq.\ref{equc} and Eq.\ref{equd}.
\begin{equation}
\hspace{-4cm}
 \begin{aligned}
    \label{equc}
   \alpha = \frac{\nu}{(1-IoU)+\nu},
\end{aligned}   
\end{equation}

\begin{equation}
\hspace{-2cm}
    \begin{aligned}
    \label{equd}
   \nu = \frac{4}{\pi^2}(arctan\frac{w^{gt}}{h^{gt}}-arctan\frac{w}{h})^2,
\end{aligned}
\end{equation}

For the training  $\lambda_{l1}=4$ and $\lambda_{iou}=2$ are used. The object query is set to $300$. Moreover, the experiment is also conducting with replacing self-supervised contrastive learning (pre-training stage) with ResNet backbone and trained using no pre-training weights and using the same parameter configuration. 

\begin{table}[b]
\centering
\caption{mAP  score on  FLIR-ADAS and KAIST Multi-spectral dataset.}
\label{table1}
\resizebox{8cm}{!}{%
\begin{tabular}{@{}l|c|c@{}}
\toprule
Dataset                                      & \multicolumn{1}{l|}{FLIR-ADAS} & \multicolumn{1}{l}{KAIST Multi-Spectral} \\ \midrule
Methods                                      & mAP Score                      & mAP Score                                \\ \midrule
Faster-RCNN +ResNet50 Backbone               & 56.70                          & 59.02                                    \\
Faster-RCNN+ ResNet50 w/ feature activation  & 62.45                          & 65.61                                    \\
Faster-RCNN +ResNet101 Backbone              & 59.03                          & 64.31                                    \\
Faster-RCNN+ ResNet101 w/ feature activation & 64.34                          & 67.52                                    \\
Encoder-decoder transformer w/ ResNet50      & 68.77                          & 62.08                                    \\
Encoder-decoder transformer w/ ResNet101     & 72.03                          & 68.25                                    \\
Self-supervised thermal network (SSTN50)     & 75.19                          & 70.62                                    \\
Self-supervised thermal network (SSTN101)    & \textbf{77.57}                 & \textbf{73.22}                           \\ \bottomrule
\end{tabular}%
}
\end{table}

\begin{table}[b]
\centering
\caption{mAP score of proposed framework with state-of-the-art methods. (*) represent average (day+night) mAP score. (-) shows that the specific algorithm is not tested on given dataset.}
\label{table2}
\resizebox{8cm}{!}{%
\begin{tabular}{@{}l|c|c@{}}
\toprule
Datasets     & \multicolumn{1}{l|}{FLIR-ADAS} & \multicolumn{1}{l}{KAIST Multi-Spectral} \\ \midrule
Methods      & mAP Score                      & mAP Score                 \\ \midrule
MMTOD-UNIT \cite{devaguptapu2019borrow}   & 61.54                          & -                         \\
MMTOD-CG \cite{devaguptapu2019borrow}     & 61.40                          & -                         \\
PiCA-Net \cite{ghose2019pedestrian}    & -                              & 65.80*                    \\
$R^3$Net \cite{ghose2019pedestrian}     & -                              & 70.85*                    \\
Intel \cite{intel}       & 23.70                          & -                         \\
tY \cite{krivsto2020thermal}          & 75.00                          & 63.00                     \\
(Our) Self-supervised Thermal Network (SSTN101) &     \textbf{77.57}                           &  \textbf{73.22}                         \\ \bottomrule
\end{tabular}%
}
\end{table}
\begin{figure}[t]
      \centering
      \includegraphics[width=8cm]{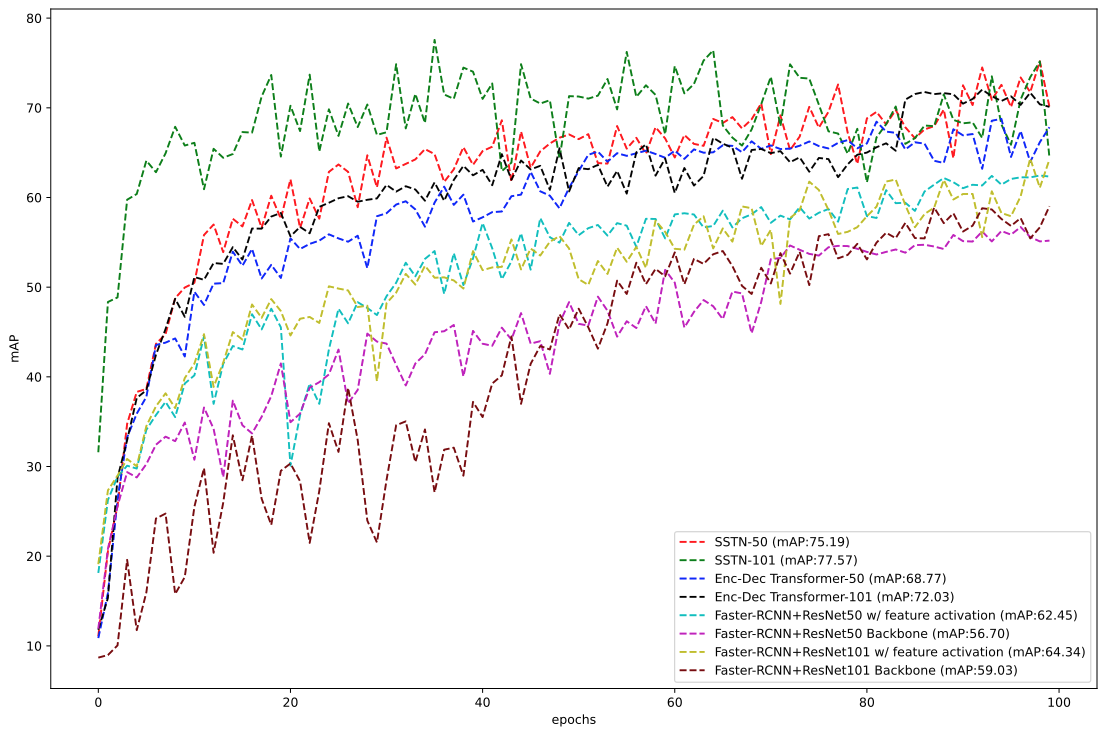}
      \caption{The illustration of mAP scores for all the models trained and tested on FLIR-ADAS dataset. The SSTN ResNet-101 has the best results.}
      \label{flir}
\end{figure}
\begin{figure}[h]
      \centering
      \includegraphics[width=8cm]{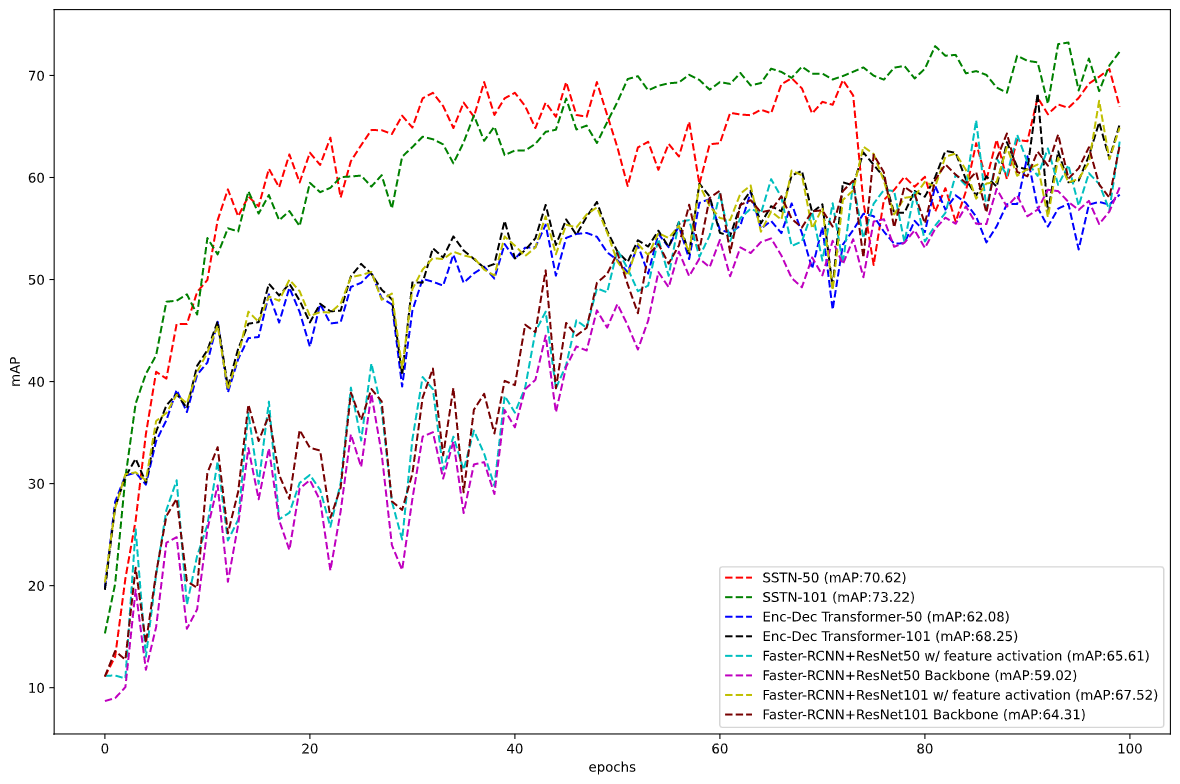}
      \caption{The illustration of mAP scores for all the models trained and tested on the KAIST Multi-Spectral dataset. The SSTN ResNet-101 has the best results.}
      \label{kiast}
\end{figure}

Table \ref{table1} shows the quantitative evaluation of the proposed algorithms. The proposed framework is evaluated on FLIR-ADAS and KAIST Multi-Spectral dataset, using ResNet 50 and ResNet101 as an encoder in the self-supervised contrastive learning stage. ResNet101 enable the self-supervised contrastive network to learn the more reliable representation of the data as the mAP score improve by 2.38\% for FLIR-ADAS and 2.6\% for KAIST Multi-Spectral dataset as compared to ResNet50. The efficacy of the self-supervised contrastive network is visible from the performance of SSTN with comparison to an encoder-decoder transformer with a ResNet backbone. An increase of 6.42\% in mAP score in FLIR-ADAS and 8.54\% mAP increase in KAIST Multi-Spectral with ResNet50 and an increase of 5.54\% mAP score in FLIR-ADAS and 4.97\% mAP improvement with ResNet101.   A similar trend is visible in the baseline consisting of the Faster-RCNN network. Using the pre-train weights to finetune on thermal images, improve the accuracy by 5.75\% in the FLIR-ADAS dataset and 6.59\% in the KAIST Multi-Spectral dataset ResNet50 and 5.31\%  increase in FLIR-ADAS and 3.21\% in KAIST Multi-Spectral with ResNet101. Fig.\ref{flir} and Fig.\ref{kiast} illustrate the quantitative comparison of mAP score between the proposed best model SSTN-101 and other variants.
\par
\begin{figure}[t]
      \centering
      \includegraphics[width=20cm,angle=90]{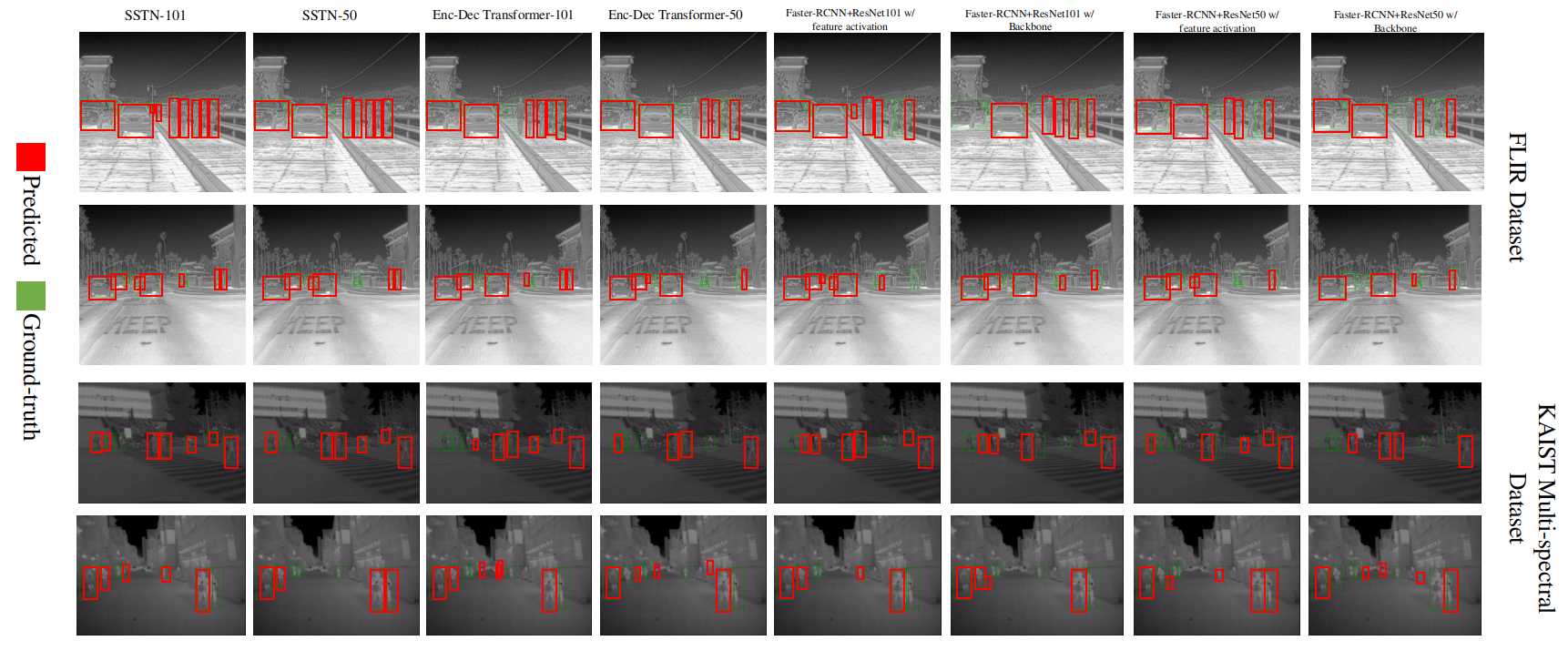}
      \caption{The qualitative results of proposed thermal object detection using FLIR-ADAS and KAIST Multi-Spectral dataset.(Best viewed in color)}
      \label{qualitative}
\end{figure}

Moreover, The proposed algorithm is compared with other-state-of-the art algorithm, as shown in Table \ref{table2}. The Self-supervised thermal network outperforms the existing algorithm by 2.57\% for FLIR-ADAS and 2.37\% for the KAIST Multi-Spectral dataset. Fig.\ref{qualitative} shows the qualitative results on FLIR-ADAS and KAIST Multi-Spectral dataset. 
 
 \section{Conclusion}
 This study focuses on thermal object detection since thermal imagery is a fundamental tool for an autonomous vehicle. We have employed the self-supervised technique to learn enhanced feature representation using unlabelled data.  A multi-scale encoder-decoder transformer network used these enhanced feature embedding to develop a robust thermal image object detector. The efficacy of a self-supervised thermal network is evaluated on FLIR-ADAS and KAIST Multi-Spectral datasets. The mean average precision of 77.57\% is achieved on the FLIR-ADAS dataset and 73.22\% is achieved on the KAIST Multi-Spectral dataset.
In future work, we aim to fuse Lidar data with thermal image data to improve object detection in adverse environmental condition. Moreover, use the thermal data to understand and classify weather condition to optimize the perception system of the autonomous vehicle. 

\section*{Acknowledgement}
This work was partly supported by the ICT R$\&$D program of MSIP/IITP (2014-0-00077, Development of global multitarget tracking and event prediction techniques based on real-time large-scale video analysis), Ministry of Culture, Sports and Tourism (MCST), and Korea Creative Content Agency (KOCCA) in the Culture Technology (CT) Research \& Development (R2020070004) Program 2021.
\bibliographystyle{IEEEtran}
\bibliography{ref.bib}

\end{document}